\documentclass{article}

    \PassOptionsToPackage{numbers, compress}{natbib}

\usepackage[preprint]{neurips_2025}

\usepackage[utf8]{inputenc} 
\usepackage[T1]{fontenc}    
\usepackage{hyperref}       
\usepackage{url}            
\usepackage{booktabs}       
\usepackage{amsfonts}       
\usepackage{nicefrac}       
\usepackage{microtype}      
\usepackage{xcolor}         
\usepackage{graphicx}       
\usepackage{booktabs}
\usepackage{subcaption}
\usepackage{listings}
\newcommand{\sys}{\textit{WGrammar}}

\title{WGRAMMAR: Leverage Prior Knowledge to Accelerate Structured Decoding}

\author{%
  Ran Wang \\
  DeepLang.ai \\
  \texttt{wrran@outlook.com}
  \And
  Xiaoxuan Liu \\
  UC Berkeley \\
  \texttt{xiaoxuan\_liu@berkeley.edu} \\
  \And
  Hao Ren \\
  DeepLang.ai \\
  \texttt{hao.ren@deeplang.ai} \\
  \AND
  Gang Chen\textsuperscript{\dag} \\
  DeepLang.ai \\
  \texttt{gang.chen@deeplang.ai} \\
  \And
  Fanchao Qi \\
  DeepLang.ai \\
  \texttt{fanchao.qi@deeplang.ai} \\
  \And
  Maosong Sun \\
  Tsinghua University \\
  \texttt{sms@tsinghua.edu.cn} \\
}

\definecolor{commentcolor}{rgb}{0.5, 0.5, 0.5}
\definecolor{stringcolor}{rgb}{0.58, 0, 0.82}

\lstdefinelanguage{FSM}{
  morekeywords=[1]{Sequence, Wait, IfElse, DoWhile},
  morekeywords=[2]{body, allow, condition, true_waits, false_waits, if_body, else_body, wait},
  sensitive=true,
  morecomment=[l]{\#},
  morestring=[b]",
}

\begin{document}

\maketitle
\footnotetext[1]{\dag: Corresponding author}

\begin{abstract}
  Structured decoding enables large language models (LLMs) to generate outputs in formats required by downstream systems, such as HTML or JSON. However, existing methods suffer from efficiency bottlenecks due to grammar compilation, state tracking, and mask creation. We observe that many real-world tasks embed strong prior knowledge about output structure. Leveraging this, we propose a decomposition of constraints into static and dynamic components—precompiling static structures offline and instantiating dynamic arguments at runtime using grammar snippets. Instead of relying on pushdown automata, we employ a compositional set of operators to model regular formats, achieving lower transition latency. We introduce \sys, a lightweight decoding engine that integrates domain-aware simplification, constraint decomposition, and mask caching, achieving up to 250$\times$ speedup over existing systems. \sys's source code is publicly available at {https://github.com/wrran/wgrammar}.
\end{abstract}

\section{Introduction}
With recent advances in large language models (LLMs), their applicability has been extended to increasingly complex tasks such as code generation~\cite{jiang2024surveylargelanguagemodels}, function calling~\cite{openai_function_calling}, and agent-based workflows~\cite{xi2023risepotentiallargelanguage, zou2025surveylargelanguagemodel, anthropic_mcp}. These tasks require structured generation to enforce a specific output format. Furthermore, in some cases, since the LLM operates as one component within a larger pipeline, its output must adhere to predefined rules to ensure that downstream components can reliably consume, parse, and build upon the model's output for continued execution.

To support structured output, users employ front-end languages to specify decoding constraints. Mainstream libraries adopt context-free grammars (CFGs) as a standard mechanism to express constraints. To efficiently enable CFG-based decoding, state-of-the-art systems~\cite{dong2024xgrammar, willard2023efficientguidedgenerationlarge} implement both a lexer and a parser to accurately interpret grammar rules. They also utilize state machines to track and enforce generation states in real time. 

Due to the inherent complexity of constrained decoding, achieving high efficiency remains a significant challenge. The overall latency can be attributed to three stages. The first is \textit{grammar compilation}, which involves lexing and parsing the front-end language to generate the internal data structures representing the constrained format. The second stage is \textit{state tracking and transition}, where the system parses each generated token and updates the state machine accordingly. Modern systems typically use pushdown automata (PDA) to support CFGs, which requires maintaining a separate stack and state machine for each request throughout its lifetime. The final stage is \textit{mask creation}, which produces a GPU-resident binary tensor of vocabulary size—where ones indicate allowed tokens and zeros indicate disallowed ones. This mask is applied to the logits prior to sampling to ensure that only valid tokens are considered during generation.
Recent efforts have sought to optimize these individual components. For example,
XGrammar~\cite{dong2024xgrammar} optimizes the automata structure to accelerate rule matching at runtime.
Despite these advances, performance remains a bottleneck.
As shown in Section~\ref{subsec:end-to-end}, for a specific workload, structured decoding introduces over $120,000$ ms of TTFT latency with Outlines and over $2,700$ ms with XGrammar. This overhead is substantial, especially given that the TTFT without structured decoding is only $525.55$ ms.

\begin{figure}[h]
    \centering
    \includegraphics[width=0.9\linewidth]{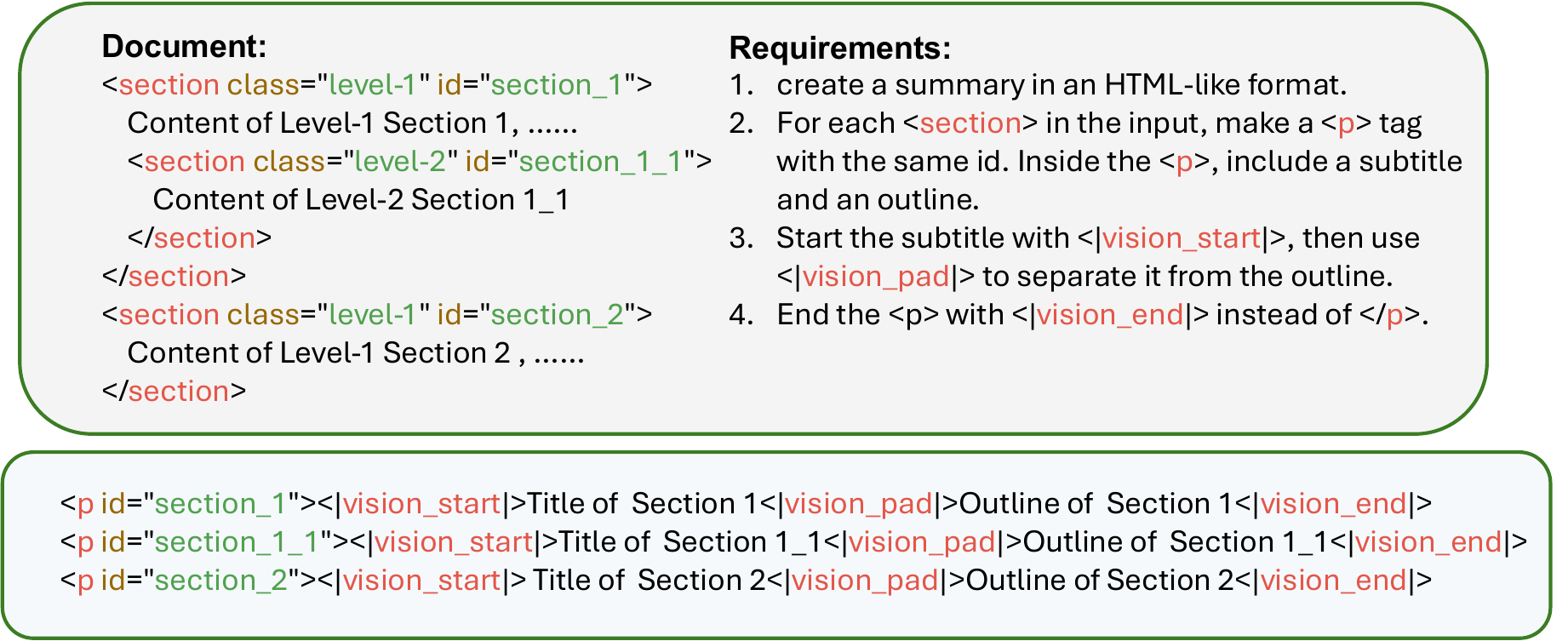}
    \caption{Structured Abstract Generation. Top: Input HTML document and requirements. Bottom: Output summary in HTML-like format.}
    \label{fig:intro-example-1}
    \includegraphics[width=0.9\linewidth]{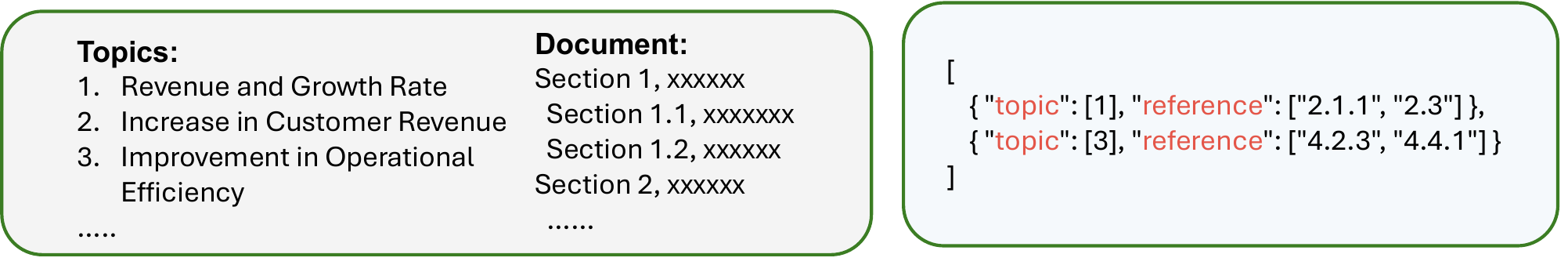}
    \caption{Reference Lookup. Left: semantic topics and document. Right: Output in JSON format, linking outline topic indices to relevant document node paths.}
    \label{fig:intro-example-2}
\end{figure}

On the other hand, a key observation overlooked by prior work is that structured decoding is often customized for highly specific downstream tasks that embed substantial prior knowledge. 
We now present two concrete examples from production workloads to illustrate what prior knowledge is and how it can be leveraged to accelerate structured decoding.
\begin{itemize}
    \item \textit{Outline Generation:} In this setting, the LLM functions as an Internet article editor. Given a raw HTML document as input, the model generates an abstract that summarizes each section while preserving the original structure. As illustrated in Figure~\ref{fig:intro-example-1}, the output must meet two requirements: (1) it includes special HTML tags to support direct visualization, and (2) it maintains the original hierarchical section structure.
    \item \textit{Reference Lookup:} As shown in Figure~\ref{fig:intro-example-2}, the LLM identifies reference paragraphs relevant to a given topic. The input is the list of topics and the document. The output is a list of JSON objects, where each object specifies a topic and the corresponding section IDs that discuss it. The output must adhere to the JSON format, using fixed keys—\texttt{"topic"} and \texttt{"reference"}—with values restricted to section IDs. 
\end{itemize}

For outline generation, the prior knowledge comes from domain-specific constraints—namely, the fact that the output will contain a fixed set of HTML tags in a predictable order. This structural knowledge allows us to precompile the format offline, rather than constructing it from scratch for each request, which would be slow due to the length and complexity of the resulting state machine (as shown in Section~\ref{chapter:parse}). 
For reference lookup, we know in advance that the output will be in a simple JSON format with specific keys and values limited to section IDs—free of deep nesting or special characters. As a result, we can predefine the transition logic from commonly used regular expressions (e.g., \texttt{\textbackslash d+}) to internal operators, and composite those basic expression snippets to the finally required state machine. 
Moreover, since we know the output does not have nesting structured, we can employ finite state machines (FSMs) instead of pushdown automata.
This optimization significantly reduces transition latency, as demonstrated in Section~\ref{chapter:exp-execution}. 
In summary, prior knowledge includes:
(1) Domain knowledge, which simplifies grammar compilation;
(2) Grammar fragments, which can be reused at runtime; and
(3) Language scope, which guides the selection of the most efficient state-tracking mechanism.

\begin{figure}[]
  \centering
  \begin{subfigure}[c]{0.45\textwidth}
    \centering
    \includegraphics[width=\textwidth]{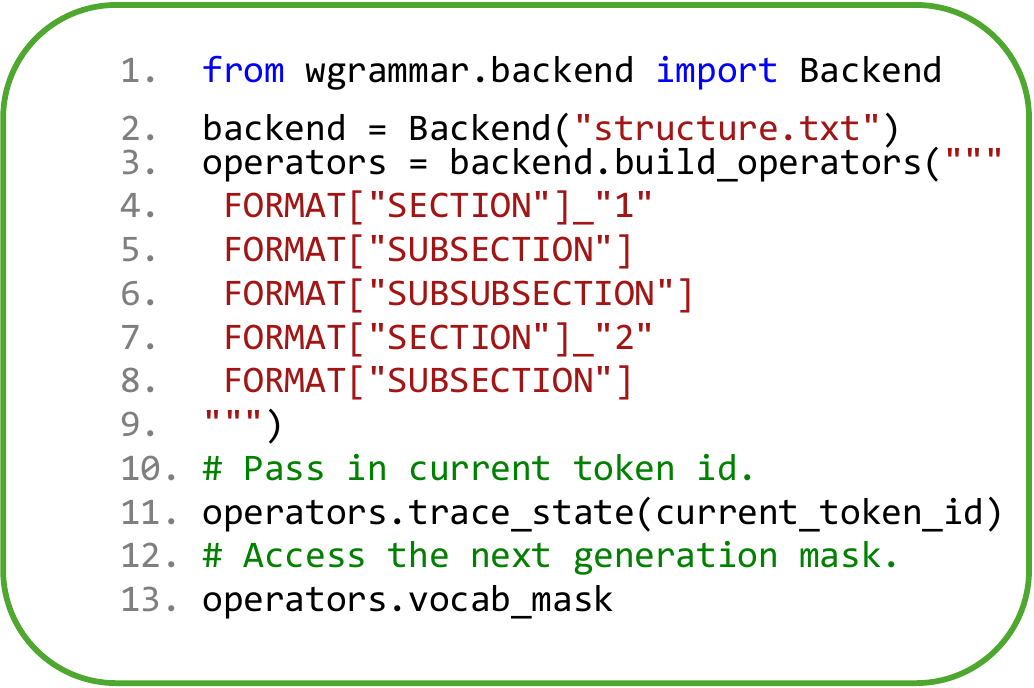}
    \caption{Example of using \sys.}
    \label{fig:api}
  \end{subfigure}
  \begin{subfigure}[c]{0.5\textwidth}
    \centering
    \includegraphics[width=\textwidth]{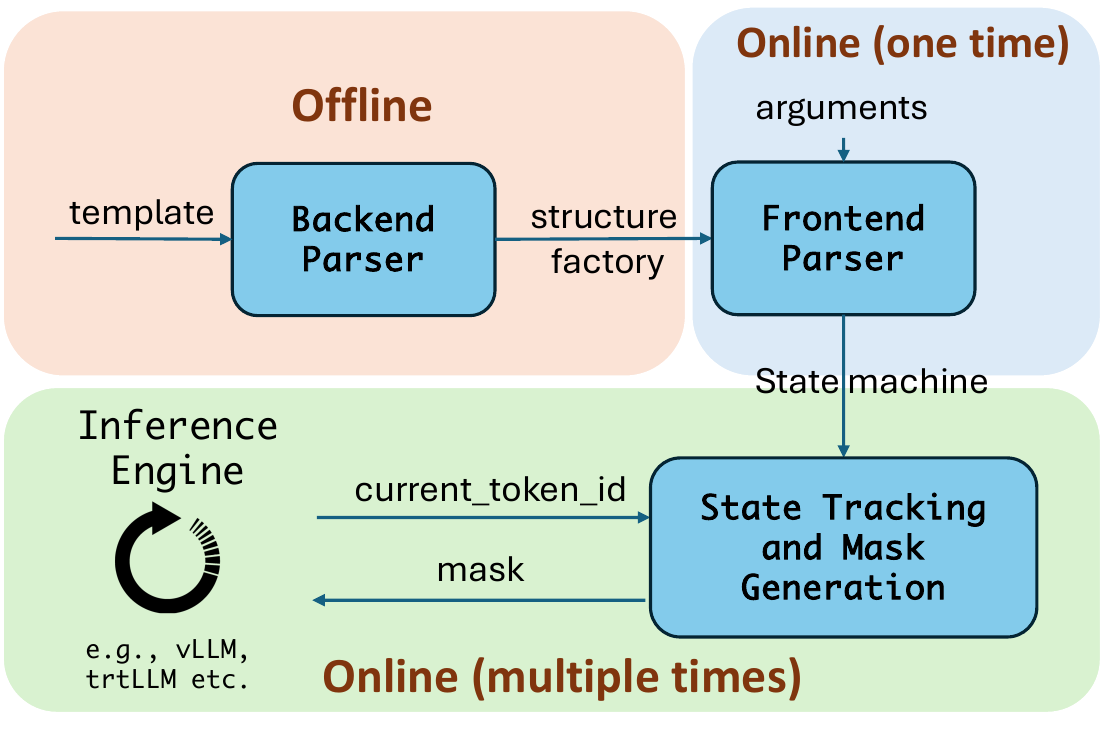}
    \caption{\sys~workflow.}
    \label{fig:sys}
  \end{subfigure}
  \label{fig:combined}
\end{figure}

Based on the observations, we build \sys~as shown in Figure~\ref{fig:sys}. \sys~consists of three main components: the backend parser, the frontend parser, and the state tracking and mask generation module. Users first define a structure template, which is sent to the backend parser to generate the structure factory. This factory contains the core structural elements—parameterized but not yet instantiated—that may be needed by incoming requests. You can think of the structure factory as a set of modular building blocks that online requests will later assemble into concrete structures. When a request arrives, its arguments are passed to the frontend parser, which uses them—along with the structure factory—to construct a state machine tailored to the request. As each token is generated, the state machine is updated accordingly to produce the appropriate mask for the next generation step. 


Figure~\ref{fig:api} shows the way of using \sys. The user initializes a \texttt{Backend} with a structure specification file (\texttt{structure.txt}), which encodes the expected structural pattern. This template includes nested section formats (e.g., SECTION, SUBSECTION, SUBSUBSECTION). 
Then a state machine is built by \texttt{build\_operators}, which tracks states and generates masks.
At each generation step, the current token ID is passed to \sys~to update the internal state machine. Based on this state, \texttt{vocab\_mask} provides the valid next-token mask that enforces the defined structure during generation.

In summary, we make the following contributions:
\begin{itemize}
    \item \textbf{Domain-aware simplification:} We observe that many structured generation tasks embed strong domain-specific constraints. By leveraging this prior knowledge, we simplify grammar design and reduce the complexity of runtime enforcement, enabling efficient, tailored decoding for specialized applications.
    \item \textbf{Constraint decomposition:} We propose a novel decomposition of constraints into static (predefined) and dynamic components. The static structure is precompiled offline into reusable templates, while the dynamic arguments are injected at runtime and might be composed using predefined grammar snippets to accelerate compilation.
    \item \textbf{Efficient implementation and release:} We implement \sys, an optimized library for constrained decoding with support for offline structure compilation, dynamic instantiation, and FSM-based tracking. \sys~outperforms the state-of-the-art systems by over 250$\times$ for TTFT and up to 2.33$\times$ for TPOT on both public benchmarks and real-world production workloads. We release all code, templates, and datasets for reproducibility.
\end{itemize}
\section{Related Work}
Structured decoding refers to techniques that constrain large language models (LLMs) to generate outputs that strictly adhere to predefined structural formats or grammars. Instead of producing free-form text, structured decoding enforces token-level constraints to ensure syntactic and semantic correctness. A common implementation strategy is to apply a dynamic mask over the model’s output logits at each generation step, allowing only tokens that are valid according to the underlying structure. This method effectively prunes invalid completions in real time, ensuring that the generated output conforms to required formats such as JSON, SQL, or domain-specific templates.

There are multiple previous works trying to support structured output. Outlines~\cite{willard2023efficientguidedgenerationlarge} first introduces a state machine to track tokens and their transition relations across generation steps, and it extends this approach by employing pushdown automata to handle context‑free grammars. SynCode~\cite{syncode} also supports CFG with a focus on on this by presenting novel parsing techniques that resolve token misalignments between the LLM’s tokenizer and structured tokens, thereby ensuring robust CFG support. XGrammar~\cite{dong2024xgrammar} further optimizes these methods by classifying tokens into context‑dependent and context‑independent categories, which significantly reduces mask‑creation time and boosts overall performance. 
On the other line, LMFE~\cite{lm-format-enforcer} narrows its focus to JSON output: by explicitly supporting required and optional JSON fields—constraints that are cumbersome to express with CFGs—it offers a lightweight, format‑guaranteed generation pipeline.

Previous works have proposed various methods to support and improve structured decoding for LLMs, yet they overlook a key observation: many production workloads incorporate substantial domain-specific knowledge, defining precise structures that can be precompiled to significantly reduce the state-machine construction overhead. Additionally, our benchmarking of state-of-the-art structured decoding libraries reveals that state transitions are on the critical path during LLM generation, incurring non-negligible runtime costs. Furthermore, since common patterns such as repetition frequently appear in many real-world workloads, we introduce some operators, such as \texttt{Wait} and \texttt{DoWhile}, to cover these basic functionalities. We employ a lightweight FSM to track the execution states of these operators, thereby substantially decreasing transition cost. Unlike pushdown automata (PDAs), whose states are contextual, our \textit{context-free} operators enable global mask caching and further improve the overall runtime efficiency.
\section{WGrammar}
In this section, we describe the major components of \sys, which include parsing, operator construction and state tracking, and mask creation. During the parsing phase, \sys~first builds an offline format factory that encodes commonly used structural patterns. To accelerate the compilation, two types of prior knowledge are incorporates: (1) domain-specific constraints, such as the requirement for certain HTML tags as shown in Figure~\ref{fig:intro-example-1}, and (2) syntax-based priors (e.g. regular expressions), like the \texttt{\textbackslash d+} pattern requiring at least one digit. At runtime, the input format is combined with these precompiled components to generate a final parse tree. This parse tree comprises operators predefined by the system, each implementing a state machine that tracks the decoding state during generation. Finally, leveraging the context-free nature of basic operators of state machines, \sys~performs global caching to efficiently create the masks required for logits processing.

\subsection{Collaborative Parsing}
\label{chapter:parse}
\begin{figure}
    \centering
    \includegraphics[width=0.99\linewidth]{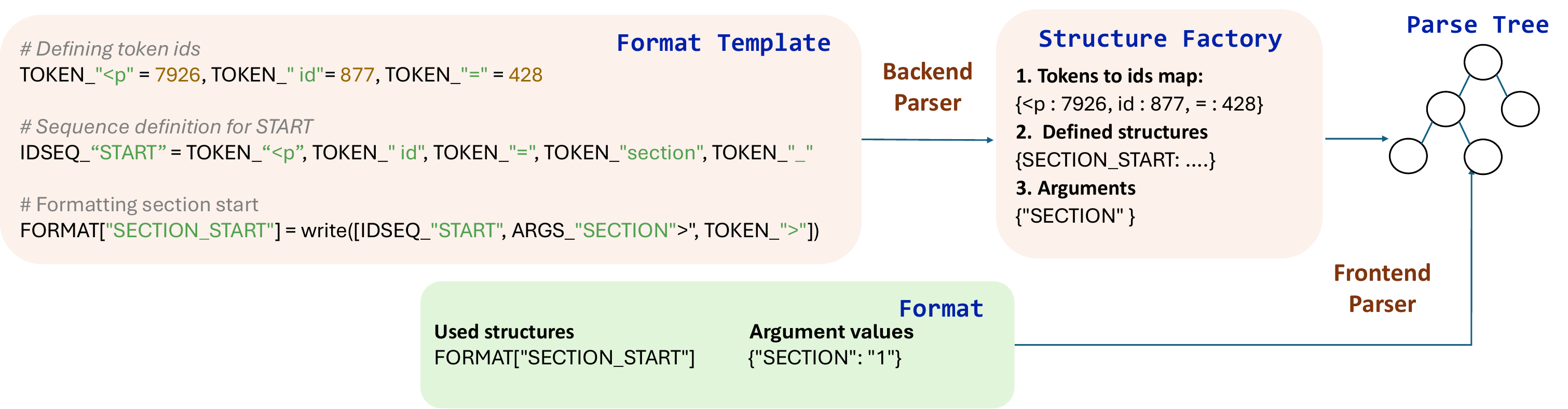}
    \caption{An example and workload of parser. The orange box shows the offline process. The green box shows the online.}
    \label{fig:parser}
\end{figure}
\sys~utilizes both offline templates and online formats to collaboratively generate parse trees. This approach is motivated by the fact that many structures in production workloads can be predetermined. Therefore, certain structures can be interpreted and stored in the structure factory. Only dynamic parts—such as the specific structure and argument values—need to be injected during runtime to generate the final parse tree. The final parse tree is then transformed to a sequence of \texttt{Wait} and \texttt{Write} operators, which will be described in greater detail in the following section.

Figure \ref{fig:parser} provides a concrete example to illustrate this workflow. In the offline phase, users define predefined format templates in EBNF~\cite{10.1145/359863.359883} grammar. For instance, the structure of \textbf{SECTION\_START} is defined in this grammar. The backend parser processes the EBNF file and generates the structure factory, which contains three key components: (1) a mapping of tokens to IDs, (2) a mapping of structure names to the corresponding data structures, and (3) the argument names that can be passed to each structure. When a request is received, it specifies the required format, including the structures and argument values. The front-end parser then combines the request format with the structure factory to generate the final parse tree. Note that, for simplicity, this example excludes any formats that cannot be represented by the offline format factory. If the input format contains elements beyond the scope of the offline factory, standard parsing based on the prior knowledge of regular expression syntax will be triggered, as described in Section~\ref{sec:op-build}.

To balance flexibility and efficiency, \sys~employ different parsing methods for offline and online phrase. The Earley algorithm~\cite{10.1145/362007.362035} is employed for the offline template parsing, whose worst-case time complexity is $O(n^3)$. Since this phase runs only one time during inference engine setup, its overhead is entirely amortized. For online formats processing, we switches to an optimized LALR(1) parser~\cite{10.1145/69622.357187}, whose linear time complexity $O(n)$ ensures runtime efficiency.

Compared to frameworks like XGrammar - which require full CFG parsing per request ($O(n^3)$ overhead) - \sys~reduces the time complexity of dynamic request processing while retaining grammatical expressiveness. Empirical results show that this design achieves a speedup of up to $251\times$ on real-world production workloads (see Section~\ref{subsec:end-to-end}).

\subsection{Operator Building and State Tracking}
\label{sec:op-build}

Domain-specific constraints usually can be directly translated into explicit operator compositions. For instance, the \texttt{SECTION\_START} in Figure~\ref{fig:parser} is mapped to a \texttt{Write} operator. The regular expressions syntax-based priors are compiled by \sys~into a compositional arrangement of operators through 
systematical translation. \sys~provides operators like \texttt{Wait}, \texttt{Write}, \texttt{IfElse}, etc., and repetition constructs (\texttt{Sequence}, \texttt{DoWhile}).
These operators collaboratively define allowable tokens, conditional checkpoints, and explicit outputs. Throughout the generation process, internal states and conditions are meticulously tracked, guiding the execution flow based on encountered tokens.

\noindent\textbf{From Regular Expressions to Wait/Write Operator Composition.}
Parsing regular expressions into structured operator compositions involves translating regex patterns into well-defined combinations of \texttt{Wait} and \texttt{Write} operators. Basic regex elements, such as character classes (\texttt{\textbackslash d}, \texttt{\textbackslash w}) or literal matches, directly map onto \texttt{Wait} operators, specifying allowable token sets. Literal sequences translate into \texttt{Write} operators, explicitly generating specific token sequences. More complex regex constructs—including repetitions (\texttt{*}, \texttt{+}), branching (\texttt{(a|b)}), and conditional patterns—are systematically transformed into structured arrangements leveraging compositional operators like \texttt{IfElse}, \texttt{Sequence}, or repetition constructs such as \texttt{DoWhile}. This translation process ensures precise control over parsing logic and token generation.
At the core of these complex structured compositions lies the ability to combine basic operators into conditional and branching structures.

\noindent\textbf{Composition of Wait/Write Operators.}
One illustrative example of this compositional capability is the \texttt{IfElse} operator, which demonstrates how the fundamental operator like \texttt{Wait} can be combined to elegantly handle conditional logic.
Internally, an \texttt{IfElse} operator is built upon a specialized \texttt{Wait} operator that differentiates conditions through token sets—\texttt{true\_waits} and \texttt{false\_waits}. Upon encountering tokens, the \texttt{IfElse} evaluates these conditions and dynamically selects the appropriate execution path (\texttt{if\_body} or \texttt{else\_body}). This design neatly captures conditional branching, enhancing basic token handling with robust, structured control-flow logic.
To fully appreciate these composite structures, it is essential to understand the fundamental operators (\texttt{Wait} and \texttt{Write}) upon which they rely.

\noindent\textbf{Wait and Write Operators.}
The foundational building blocks of structured token management are the \texttt{Wait} and \texttt{Write} operators. The \texttt{Wait} operator functions as a conditional checkpoint within token sequences, defining specific tokens (\texttt{waits}) that signal pauses or conditional evaluations, optionally constraining permissible tokens using additional sets (\texttt{denies} or \texttt{allows}). Upon satisfying a defined condition (matching a token in \texttt{waits}), it optionally executes nested operators (\texttt{body}). In contrast, the \texttt{Write} operator straightforwardly outputs predetermined token sequences. Together, these basic operators establish the groundwork for complex structured token parsing and generation processes, enabling precise control over token transitions and output generation.

Lastly, we use a concrete example to show how a regular expression is translated into building operators.

\begin{minipage}[t]{0.55\textwidth}
\begin{lstlisting}
Sequence(
  Wait(allow={\d}, wait={\d}),
  IfElse(
    condition=Wait(allow={\d, \., <eos>},  
                   true_waits={<eos>},    
                   false_waits={\.}),     
    if_body=None,
    else_body=DoWhile(
        body=Wait(allow={\d}, wait={\d}),
        condition=Wait(allow={\d, \., <eos>},
                       true_waits={\.},
                       false_waits={<eos>})
    )
  )
)
\end{lstlisting}
\end{minipage}
\hspace{1em}
\begin{minipage}[t]{0.42\textwidth}
The left code structure represents the regular expression pattern \texttt{\textbackslash d+(\textbackslash.\textbackslash d+)*}. This structure is implemented by first using a \texttt{Wait} operator to enforce the generation of at least one digit, corresponding to the 
\texttt{\textbackslash d+} prefix. Following this, an \texttt{IfElse} construct examines whether the next token is a digit, a period, or an end-of-sequence marker. If the next token is <eos>, the sequence terminates, representing the case where only a single digit sequence is present. If the next token is a period, the grammar enters a \texttt{DoWhile} loop that repeatedly matches additional digit sequences, each introduced by a period.
\end{minipage}

The loop continues as long as a period is encountered, and terminates cleanly upon seeing <eos>, ensuring the output conforms precisely to the expected hierarchical format. This decomposition enables the generation process to enforce structural constraints at each step, ensuring the output adheres strictly to the intended pattern without requiring post hoc validation.

\subsection{Mask Generation and Global Caching}
The base operators, \texttt{Wait} and \texttt{Write}, are responsible for the final generation of the logit mask. The \texttt{Write} operator can be viewed as a sequence of specialized \texttt{Wait} operators, whose element explicitly specifies the expected token. As mentioned above, the \texttt{Wait} operator can exclude invalid tokens through a set \texttt{denies}, or specify valid tokens through a set \texttt{allows}. We store a few invalid tokens in the \texttt{denies} in accept-heavy cases; conversely, in reject-heavy cases, the \texttt{allows} can be employed to store the accepted tokens. The adaptive storage mechanism reduces the cost involved in constructing the fresh mask, thereby enhancing efficiency. 

This modular and deterministic nature of mask generation also lays the foundation for further optimization. Specifically, our final state machine is constructed using boolean-capable \texttt{Wait} operators (through \texttt{true\_waits} and \texttt{false\_waits}) along with essential compositional operators such as \texttt{Sequence}, \texttt{IfElse}, and \texttt{DoWhile} for state tracking. Unlike PDAs, which rely on \textit{contextual} states to resolve ambiguities, these operators are \textit{context-free} in nature and generate the mask based solely on the \texttt{allows} or \texttt{denies}. This design enables further optimization through the use of a \textbf{global cache}, which amortizes the cost of mask construction across multiple requests and decoding steps.
\section{Experiments}
\label{sec:experiment}
In this section, we first evaluate the end-to-end latency of \sys. We then break down the execution time and analyze the time spent in each stage.

\subsection{Experiment Setup}
We implement the \sys~framework in Python and integrate it with vLLM-v1~\cite{vllm} (v0.8.5.post1), a high-performance inference library for LLMs.

\noindent{\textbf{Baselines}}
To evaluate the performance of our proposed method, we compare it with two state-of-the-art structured generation methods:
\begin{itemize}
\item Outlines (v0.2.3)~\cite{willard2023efficientguidedgenerationlarge}: A model-agnostic approach that reformulates text generation as transitions between states in finite-state machine with minimal overhead during token generation.
\item 
XGrammar (v0.1.18)~\cite{dong2024xgrammar}: An efficient grammar-guided decoding engine that accelerates generation by separating vocabulary tokens into context-independent and context-dependent subsets.
\end{itemize}

These baselines are chosen due to their proven effectiveness in structured text generation tasks and their availability for benchmarking purposes.
To examine the performance of \sys~ without relying on domain-specific priors, we also include a variant of \sys~ that only leverages syntax-based priors derived from regular expressions as a competitive baseline, referred to as \texttt{WGrammar (online)}.

\noindent{\textbf{Datasets and Tasks}}
We conduct experiments on three distinct datasets:
\begin{itemize}
\item Outline-Generation: This dataset requires the LLM to generate outline corresponding to the original input. The output should consist of special HTML tags and summaries, where the structure must accurately reflect the organization of the source material.
\item Reference-Lookup: In this dataset, the LLM is tasked with identifying relevant paragraphs for a given topic and returning them in a JSON-formatted array. Each element in the array includes a \texttt{topic} key specifying the main theme and a \texttt{reference} key containing related paragraphs.
\item JSON-mode-eval~\cite{json-mode-eval}: For this dataset, the prompt explicitly provides a JSON schema that defines the expected output structure. The LLM is required to generate content that conforms to this schema base on the provided context.
\end{itemize}


\noindent{\textbf{Model and Hardware}}
In this study, we employed different versions of the Qwen2.5 model~\cite{qwen2.5} to suit the task requirements of each dataset. Specifically, for the Outline-Generation and Reference-Lookup datasets, we used a 7B version of Qwen2.5. The models for these tasks were fine-tuned to enhance their instruction-following capabilities. For the JSON-mode-eval dataset, a smaller 0.5B parameter version of Qwen2.5 was adopted, which can strictly adhere to the provided JSON schema during text generation.

All experiments were conducted on systems equipped with NVIDIA A100 GPUs to ensure consistent and high-performance computing resources. To eliminate potential biases introduced by parallel request processing in the inference engine, we processed each instance sequentially. This setup allows for a clearer comparison of the performance differences among the structured generation frameworks, without being obscured by concurrent execution mechanisms.

\subsection{End to End Results}
\label{subsec:end-to-end}

\begin{table}[htbp]
\centering
\caption{Overhead comparison of methods across three datasets. The TTFT w/o structured generation on the three datasets are $525.55$ms, $104.87$ms and $18.66$ms respectively. The TPOTs are $11.71$ms, $11.32$ms and $3.69$ms respectively. Numbers in brackets show the reduction compared with Outlines.}
\label{tab:overhead}
\begin{tabular}{crrrr}
\toprule
\textbf{Dataset} & \multicolumn{1}{c}{\textbf{Outlines}} & \multicolumn{1}{c}{\textbf{XGrammar}} & \multicolumn{1}{c}{\textbf{WGrammar (online)}} & \multicolumn{1}{c}{\textbf{WGrammar}} \\
\midrule
\multicolumn{5}{c}{\texttt{Time to First Token overhead}} \\
\midrule
Outlines-Generation & $120,234.58$ & $2,747.83$ ($44\times$) & $64.25$ ($1,871\times$) & $10.93$ ($11,000\times$) \\
Reference-Lookup & $3,174.34$ & $116.89$ ($27\times$) & $4.17$ ($761\times$) & $4.70$ ($675\times$) \\
JSON-mode-eval & $6,164.39$ & $137.68$ ($45\times$) & $7.56$ ($815\times$) & $1.38$ ($4,467\times$)\\
\midrule
\multicolumn{5}{c}{\texttt{Time per Output Token overhead}} \\
\midrule
Outlines-Generation & $24.02$ & $1.61$ ($15\times$) & $0.93$ ($26\times$) & $0.69$ ($35\times$) \\
Reference-Lookup & $21.64$ & $0.51$ ($42\times$) & $0.42$ ($52\times$) & $0.50$ ($43\times$) \\
JSON-mode-eval & $20.55$ & $0.89$ ($25\times$) & $0.70$ ($32\times$) & $0.67$ ($34\times$) \\
\bottomrule
\end{tabular}
\end{table}

To evaluate the practical performance of various methods, we focused on two key metrics: TTFT (time to first token), which measures the latency from the start of the request to the generation of the first token, and TPOT (time per output token), which reflects the average time taken to generate each subsequent token after the first one. The metric TTFT reflects the cost of constructing the automata; while the metric TPOT captures the decoding efficiency during the generation process.
We first present the overheads for TTFT and TPOT, defined as the additional latency introduced compared to decoding without structured generation. As shown in Table~\ref{tab:overhead}, \sys~reduces the structured decoding overhead by over 250$\times$ for TTFT and up to 2.33$\times$ for TPOT compared to XGrammar.

\begin{figure}[htb]
\centering
\begin{subfigure}[b]{0.99\textwidth}
\includegraphics[width=\linewidth]{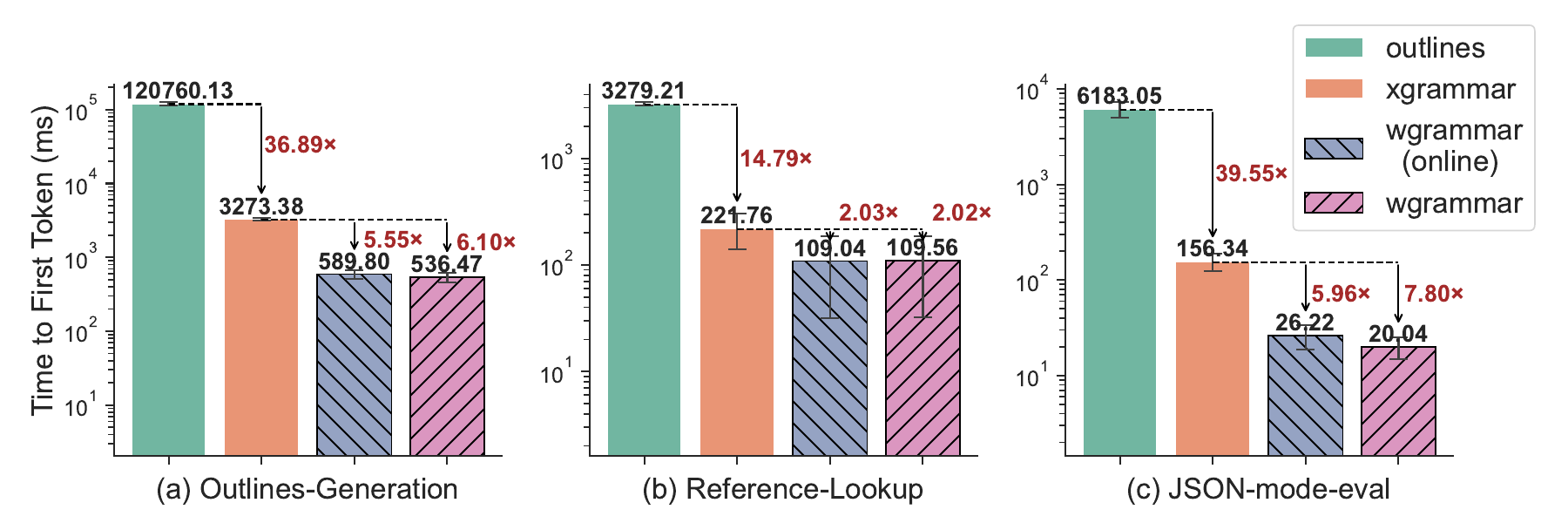}
\caption{TTFT (time to first token) across different datasets.}
\label{fig:ttft}
\end{subfigure}

\centering
\begin{subfigure}[b]{0.99\textwidth}
\includegraphics[width=0.99\linewidth]{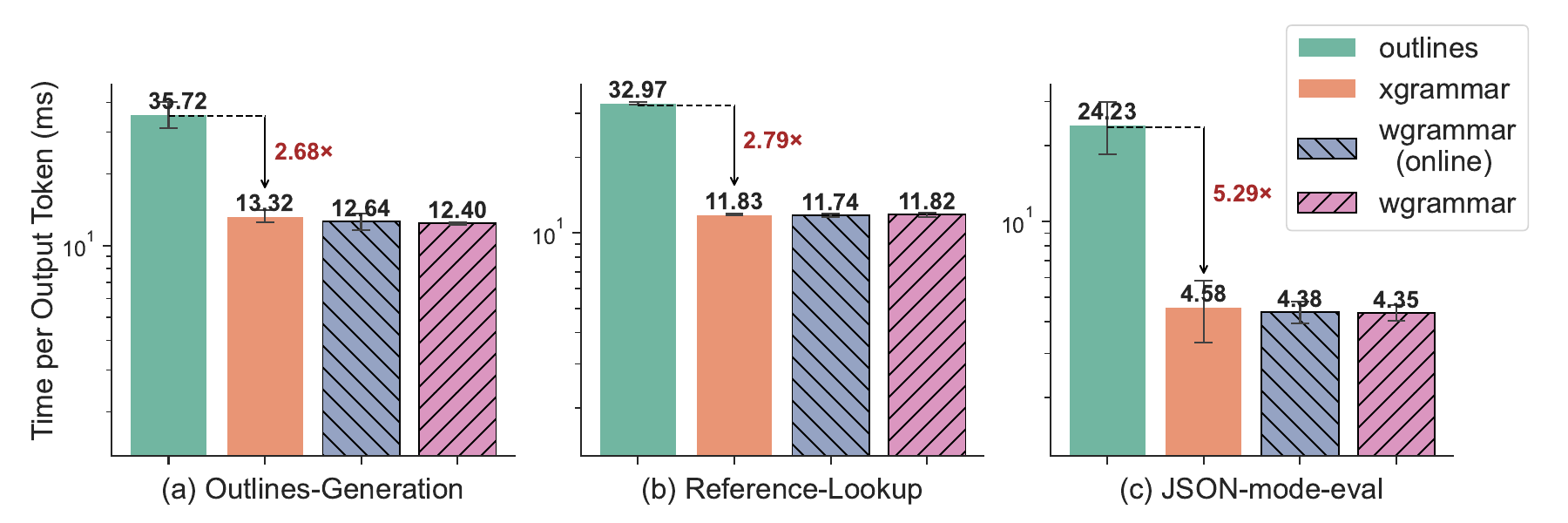}
\caption{TPOT (time per output token) across different datasets.}
\label{fig:tpot}
\end{subfigure}
\caption{End-to-end experiment results. The y-axis is in log scale for demonstration.}
\label{fig:main-result}
\end{figure}

Next, we show the end-to-end TTFT and TPOT numbers in Figure~\ref{fig:main-result}. \sys~achieves the best TTFT and TPOT among all baselines for all three datasets. 

Figure~\ref{fig:ttft} shows that, across all datasets, the TTFT of \sys~is the smallest. Particularly in the structurally complex Outline-Generation dataset, \sys ~achieves a speedup of $225\times$ compared to Outlines. Compared to XGrammar, \sys~still offers an improvement of $6.10\times$.
This is because the output structure of Outlines-Generation is highly complex. When using XGrammar, it requires more than $60$ rules to fully describe the final structure, which imposes a non-negligible cost of compilating the PDAs; in contrast, with \sys, only a few arguments need to be specified to quickly construct the state machine.
Even when using only regular expression syntax priors, it achieves a $5.55\times$ speedup in TTFT overhead compared to XGrammar. On other datasets, \sys~also achieves acceleration ranging from $2.02\times$ to $7.80\times$. This demonstrates the effectiveness of the collaborative parsing strategy in \sys.

Figure~\ref{fig:tpot} illustrates the time for generating each token. Similarly, \sys~exhibits the lowest overhead across all datasets. Thanks to its low time complexity state tracking and global caching mechanisms, \sys's advantages become even more pronounced when generating longer texts as on the Outlines-Generation dataset. It reduces the overhead of the carefully designed XGrammar from $13.32$ ms to $12.40$ ms. This highlights the benefits of \sys~in long-running systems.

\subsection{Execution Overhead Breakdown}
\label{chapter:exp-execution}
To better understand the performance characteristics of structured generation methods, we conduct a fine-grained analysis of the execution overhead breakdown across the three key stages: \textit{grammar compilation}, \textit{state tracking and transition}, and \textit{mask creation}.

\begin{figure}
    \centering
    \includegraphics[width=0.99\linewidth]{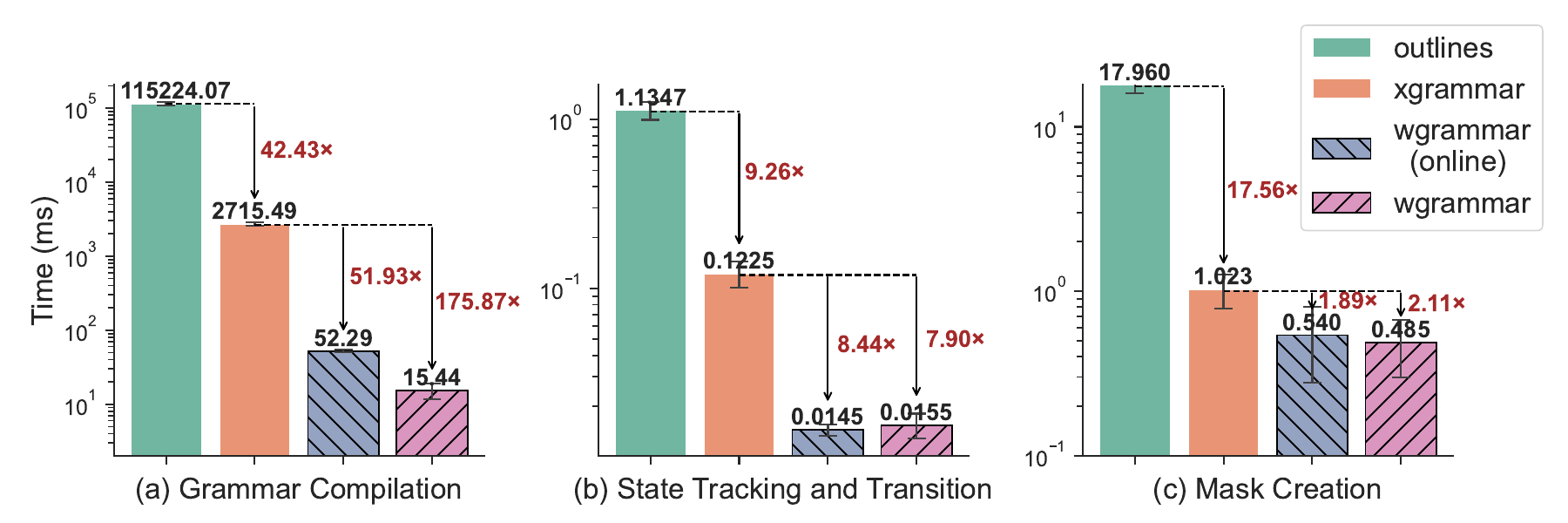}
    \caption{Execution overhead breakdown of different methods on the Outlines-Generation dataset. WGrammar demonstrates superior performance across all three stages.}
    \label{fig:breakdown}
\end{figure}

Figure~\ref{fig:breakdown} shows all methods' overhead breakdown on the Outlines-Generation dataset. 
The superior performance of \sys~in the \textit{grammar compilation} stage—outperforming the current state-of-the-art framework, XGrammar, by a factor of $175\times$—can be attributed to its collaborative parsing strategy. These findings are consistent with our previous observations regarding TTFT in Figure~\ref{fig:main-result}.

Similarly, the cost of \textit{state tracking and transition} is significantly lower in real production environments. Specifically, WGrammar's Python implementation outperforms XGrammar's C++ implementation by a factor of $7\times$ to $8\times$.
This advantage is attributed to \sys~providing a set of highly efficient operators  such as \texttt{IfElse}, \texttt{DoWhile}, etc., where state tracking requires only constant time complexity. In constrast, XGrammar must maintain the contextual states of PDAs, which incurs significant overhead even using a high-performance implementation.
This efficiency gain underscores the effectiveness of WGrammar's optimized state management mechanisms.

Regarding \textit{mask creation}, although WGrammar does not utilize more compact and efficient data structures as XGrammar, it benefits from the global caching mechanism. As the system runs longer, the amortized complexity of mask creation decreases. Consequently, WGrammar still achieves superior performance in this phase, with an approximate speedup of 2$\times$ over other methods.

\section{Conclusion}

We presented \sys, a structured decoding engine that leverage prior knowledge to decompose output constraints into static and dynamic components. By combining offline compilation with efficient runtime instantiation and using FSMs based on various operators in place of PDAs, \sys~significantly reduces the overhead of grammar compilation, state tracking and mask generation. Experimental results show that \sys~achieves up to 250$\times$ over XGrammar, enabling efficient and flexible deployment of LLMs in format-sensitive applications.

\section{Limitations and Broader Impact}
\label{sec:limitations}
By enabling precise control over output structures while maintaining high performance, \sys~advances the practical deployment of structured generation in real-world applications. But additional work is needed to manually design the offline templates to achieve optimal performance, which introduces challenges for users unfamiliar with the backend DSLs. Instead, those unfamiliar with DSLs can use regular expressions, which WGrammar supports, to describe structural constraints. Another limitations is that \sys~uses \textit{non-greedy matching} for regular expressions, differing from the more common greedy matching in text processing. This distinction requires users to understand these semantic differences.

Our work introduces an efficient and production-ready framework for structured generation. The positive impact lies in providing a practical solution that supports complex output structures - range from JSON to HTML-like markup - without compromising inference efficiency. This advancement lays a solid foundation for building robust systems that rely on structured LLM outputs.

\bibliographystyle{plain}
\bibliography{ref}

\begin{thebibliography}{10}

\bibitem{anthropic_mcp}
{Anthropic}.
\newblock Introducing the model context protocol.
\newblock \url{https://www.anthropic.com/news/model-context-protocol}, 2024.
\newblock Accessed: 2025-05-11.

\bibitem{10.1145/69622.357187}
Frank DeRemer and Thomas Pennello.
\newblock Efficient computation of lalr(1) look-ahead sets.
\newblock {\em ACM Trans. Program. Lang. Syst.}, 4(4):615–649, October 1982.

\bibitem{dong2024xgrammar}
Yixin Dong, Charlie~F Ruan, Yaxing Cai, Ruihang Lai, Ziyi Xu, Yilong Zhao, and Tianqi Chen.
\newblock Xgrammar: Flexible and efficient structured generation engine for large language models.
\newblock {\em arXiv preprint arXiv:2411.15100}, 2024.

\bibitem{10.1145/362007.362035}
Jay Earley.
\newblock An efficient context-free parsing algorithm.
\newblock {\em Commun. ACM}, 13(2):94–102, February 1970.

\bibitem{lm-format-enforcer}
Noam Gat.
\newblock {lm-format-enforcer}.
\newblock \url{https://github.com/noamgat/lm-format-enforcer}, 2023.
\newblock Accessed: 2024-05-09.

\bibitem{jiang2024surveylargelanguagemodels}
Juyong Jiang, Fan Wang, Jiasi Shen, Sungju Kim, and Sunghun Kim.
\newblock A survey on large language models for code generation, 2024.

\bibitem{vllm}
Woosuk Kwon, Zhuohan Li, Siyuan Zhuang, Ying Sheng, Lianmin Zheng, Cody~Hao Yu, Joseph Gonzalez, Hao Zhang, and Ion Stoica.
\newblock Efficient memory management for large language model serving with pagedattention.
\newblock In Jason Flinn, Margo~I. Seltzer, Peter Druschel, Antoine Kaufmann, and Jonathan Mace, editors, {\em Proceedings of the 29th Symposium on Operating Systems Principles, {SOSP} 2023, Koblenz, Germany, October 23-26, 2023}, pages 611--626. {ACM}, 2023.

\bibitem{json-mode-eval}
NousResearch.
\newblock json-mode-eval dataset, 2023.

\bibitem{openai_function_calling}
{OpenAI}.
\newblock Function calling guide.
\newblock \url{https://platform.openai.com/docs/guides/function-calling?api-mode=responses}, 2024.
\newblock Accessed: 2025-05-11.

\bibitem{syncode}
Shubham Ugare, Tarun Suresh, Hangoo Kang, Sasa Misailovic, and Gagandeep Singh.
\newblock Syncode: Llm generation with grammar augmentation, 2024.

\bibitem{willard2023efficientguidedgenerationlarge}
Brandon~T. Willard and Rémi Louf.
\newblock Efficient guided generation for large language models, 2023.

\bibitem{10.1145/359863.359883}
Niklaus Wirth.
\newblock What can we do about the unnecessary diversity of notation for syntactic definitions?
\newblock {\em Commun. ACM}, 20(11):822–823, November 1977.

\bibitem{xi2023risepotentiallargelanguage}
Zhiheng Xi, Wenxiang Chen, Xin Guo, Wei He, Yiwen Ding, Boyang Hong, Ming Zhang, Junzhe Wang, Senjie Jin, Enyu Zhou, Rui Zheng, Xiaoran Fan, Xiao Wang, Limao Xiong, Yuhao Zhou, Weiran Wang, Changhao Jiang, Yicheng Zou, Xiangyang Liu, Zhangyue Yin, Shihan Dou, Rongxiang Weng, Wensen Cheng, Qi~Zhang, Wenjuan Qin, Yongyan Zheng, Xipeng Qiu, Xuanjing Huang, and Tao Gui.
\newblock The rise and potential of large language model based agents: A survey, 2023.

\bibitem{qwen2.5}
An~Yang, Baosong Yang, Beichen Zhang, Binyuan Hui, Bo~Zheng, Bowen Yu, Chengyuan Li, Dayiheng Liu, Fei Huang, Haoran Wei, Huan Lin, Jian Yang, Jianhong Tu, Jianwei Zhang, Jianxin Yang, Jiaxi Yang, Jingren Zhou, Junyang Lin, Kai Dang, Keming Lu, Keqin Bao, Kexin Yang, Le~Yu, Mei Li, Mingfeng Xue, Pei Zhang, Qin Zhu, Rui Men, Runji Lin, Tianhao Li, Tingyu Xia, Xingzhang Ren, Xuancheng Ren, Yang Fan, Yang Su, Yichang Zhang, Yu~Wan, Yuqiong Liu, Zeyu Cui, Zhenru Zhang, and Zihan Qiu.
\newblock Qwen2.5 technical report.
\newblock {\em CoRR}, abs/2412.15115, 2024.

\bibitem{zou2025surveylargelanguagemodel}
Henry~Peng Zou, Wei-Chieh Huang, Yaozu Wu, Yankai Chen, Chunyu Miao, Hoang Nguyen, Yue Zhou, Weizhi Zhang, Liancheng Fang, Langzhou He, Yangning Li, Yuwei Cao, Dongyuan Li, Renhe Jiang, and Philip~S. Yu.
\newblock A survey on large language model based human-agent systems, 2025.

\end{thebibliography}

\end{document}